\documentclass[letterpaper, 10 pt, conference]{ieeeconf}  

\IEEEoverridecommandlockouts                              

\usepackage{graphics} 
\usepackage{epsfig} 
\usepackage{mathptmx} 
\usepackage{times} 
\usepackage{amsmath} 
\usepackage{amssymb}  
\usepackage{float}
\usepackage{threeparttable}
\usepackage{textcomp}
\usepackage{siunitx}
\usepackage{adjustbox}
\usepackage{diagbox}
\usepackage{booktabs, multirow}
\usepackage[utf8]{inputenc}
\usepackage{xcolor}
\usepackage{graphicx}
\usepackage{tikz}
\usepackage{listings}                       
\usetikzlibrary{backgrounds}

\usepackage[noadjust]{cite}
\usepackage{gensymb}
\usepackage{algorithm}
\usepackage[noend]{algpseudocode}
\usepackage[table]{xcolor}
\usepackage{pifont}
\newcommand{\cmark}{\ding{51}}
\newcommand{\xmark}{\ding{55}}
\newcommand{\cv}[1]{\textcolor{black!55}{#1}} 
\usepackage{placeins}
\usepackage{microtype}

\title{\LARGE \bf
Scene-Q: Confidence-Aware Coarse-to-Fine Querying of 3D Scenes with Selective VLM Reasoning}

\author{Juno Kim$^{1*}$ Yesol Park$^{1*}$ Hye-Jung Yoon$^{1*}$ Byoung-Tak Zhang$^{1,2,3}$%
    \thanks{*Authors have equal contributions}
    \thanks{$^{1}$Interdisciplinary Program in AI, Seoul National University}%
    \thanks{$^{2}$Artificial Intelligence Institute, Seoul National University}%
    \thanks{$^{3}$Department of Computer Science, Seoul National University}%
    \thanks{This work was partly supported by Institute of Information \& communications Technology Planning \& Evaluation (IITP) grant funded by the Korea government(MSIT) [RS-2021-II211343, Artificial Intelligence Graduate School Program (Seoul National University)] and (RS-2021-II212068-AIHub/10\%, RS-2021-II211343-GSAI/15\%, 2022-0-00951-LBA/15\%, 2022-0-00953-PICA/20\%), NRF (RS-2024-00353991/20\%, RS-2023-00274280/10\%), and KEIT (RS-2024-00423940/10\%) grant funded by the Korean government.
    }%
}

\begin{document}
\maketitle

\thispagestyle{empty}
\pagestyle{empty}


\begin{abstract}
Indoor mobile robots require open-vocabulary scene understanding that grounds natural-language queries in a consistent 3D map. Many existing systems ultimately rely on cosine-similarity retrieval with contrastive image--text encoders, which is efficient but brittle when labels are near-synonymous or multiple similar instances appear. We present Scene-Q, a confidence-aware coarse-to-fine querying framework that normalizes encoder scores with temperature scaling and selectively invokes a reasoning VLM only for low-confidence cases. High-confidence queries are answered by fast retrieval, while ambiguous ones are reranked over a small top-$K$ candidate set using the original multi-view images and instance bounding boxes, enabling context-aware disambiguation at low cost. Scene-Q improves open-vocabulary 3D instance segmentation on ScanNet200 and natural-language 3D instance retrieval on real-world reconstructions, with the largest gains on spatial and relational queries while keeping a substantial fraction of queries on the fast path.
\end{abstract}

\section{INTRODUCTION}
Indoor mobile robots increasingly operate in real-world homes and offices, where users expect open-ended natural-language queries (e.g., object identity, attributes, and spatial relations) to be grounded in a consistent 3D map. Recent systems combine 3D mapping with vision--language foundation models for open-vocabulary scene understanding~\cite{takmaz2023openmask3d, kim2024ov, nguyen2024open3dis}. In practice, however, many pipelines still make the final decision with contrastive image--text encoders~\cite{radford2021learningCLIP, sun2023eva, yu2022coca, zhai2023sigmoid, tschannen2025siglip} via cosine-similarity matching, which is efficient but brittle under ambiguity and weak at context-sensitive, compositional reasoning. This brittleness becomes pronounced for near-synonymous labels and crowded scenes, where global context and inter-object relations are crucial.

\begin{figure} [t!]
\begin{center}
\includegraphics[width=0.9\columnwidth]{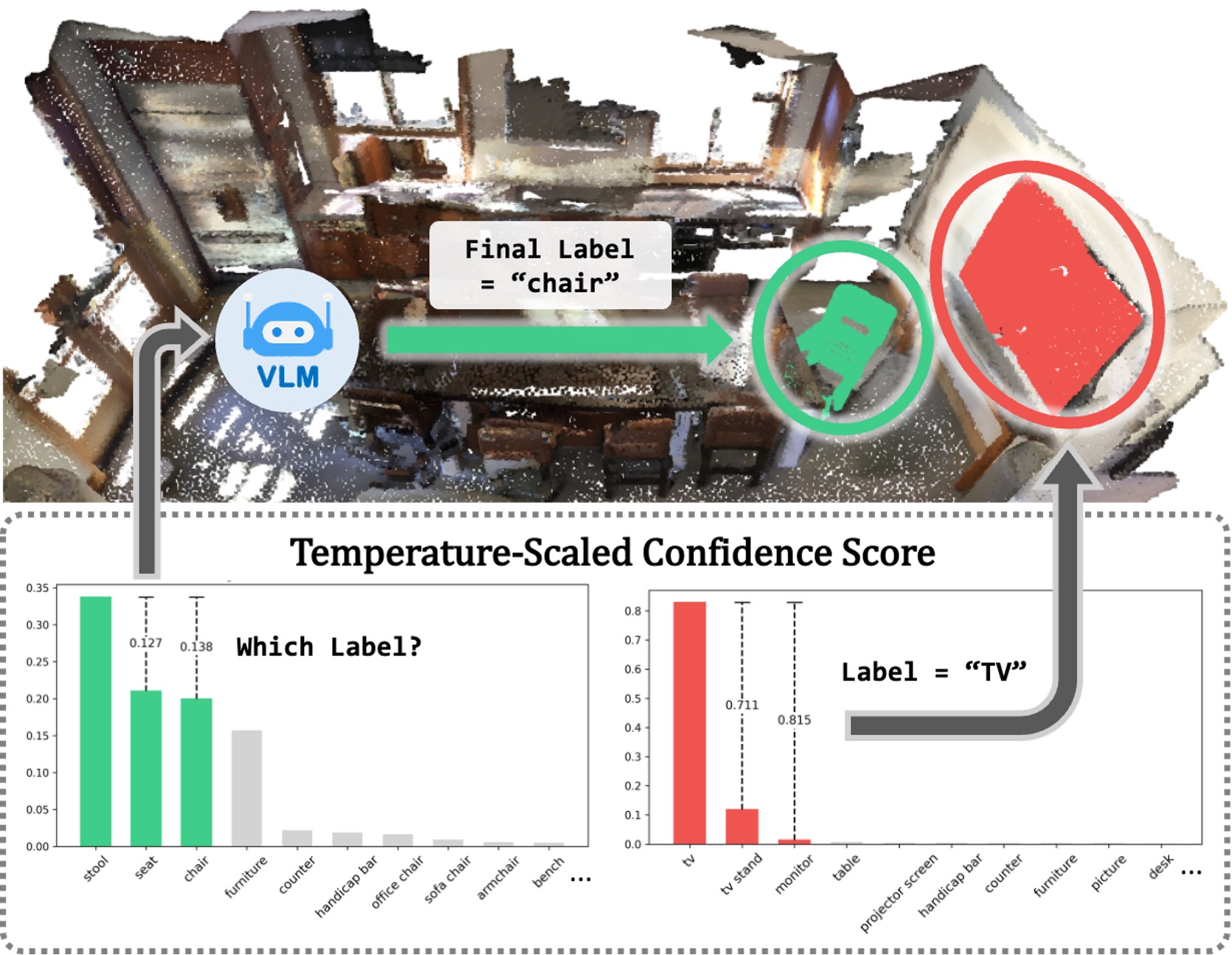}
\vspace{-2mm}
\caption{\textbf{Selective VLM Reasoning for Scene-Q.} The figure shows a 3D point-cloud scene containing two masked instances (chair and TV). 
For each instance, we show temperature-scaled image–text scores over ScanNet200 labels. 
The TV yields a peaked, high-confidence distribution (e.g., \texttt{tv} $\approx 0.83$), so the fast encoder prediction is selected. 
The chair produces tightly clustered scores across near-synonyms (e.g., \texttt{stool}/\texttt{seat}/\texttt{chair}/\dots); low-confidence cases form the top-$K$ candidates and consult a reasoning VLM with full-image context and a bounding box to select the correct label. 
High-confidence cases follow the fast path, while low-confidence cases are routed to a reasoning VLM for further disambiguation.}
\label{system_proposal}
\vspace{-8mm}
\end{center}
\end{figure}

We observe two recurring failure modes in image–text–encoder–based open-vocabulary 3D instance segmentation and querying. First, in \emph{instance$\rightarrow$texts} label assignment, as in ScanNet200~\cite{dai2017scannet, rozenberszki2022language}, systems compare one 3D instance against hundreds of text labels via cosine similarity. Near-synonymous or fine-grained labels—chair, armchair, folding chair, stool—produce tightly clustered scores, yielding unstable top-1 choices even when the instance is unambiguous. Second, in \emph{text$\rightarrow$instances} retrieval with descriptive queries, when multiple similar instances exist in a scene, cosine scores again bunch together and the top-1 often selects the wrong target. In practice, to obtain instance-level features with image--text encoders that produce a single global image representation, pipelines typically crop the instance in 2D views before encoding. This crop-based practice discards global cues—room type, object relations, scene layout—that are crucial for disambiguating fine-grained categories and resolving multi-instance ambiguity, reinforcing the weaknesses above.

\begin{figure*} [t!]
\begin{center}
\vspace{2mm}
\includegraphics[width=0.91\textwidth]{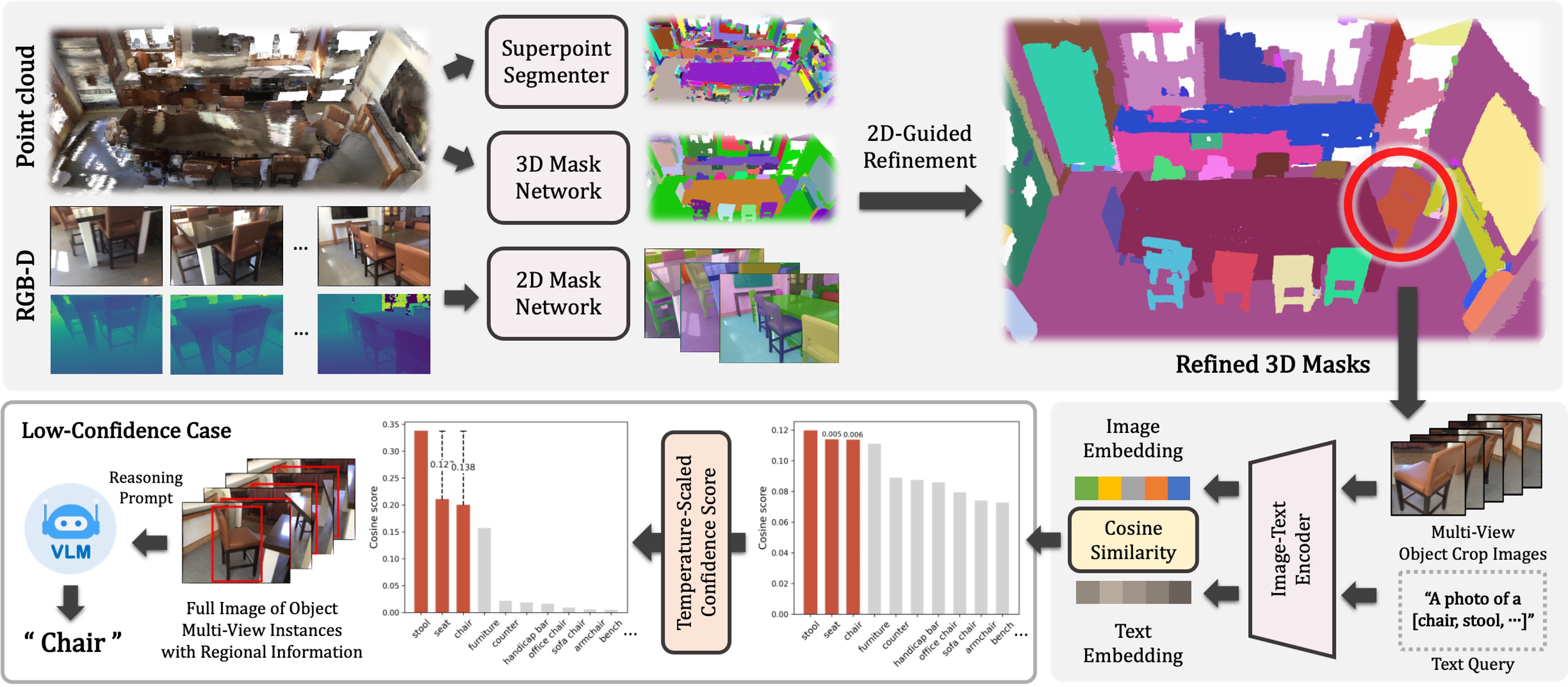}
\vspace{-2mm}
\caption{\textbf{Overview of Scene-Q.} Our pipeline has three stages: (1) From registered RGB-D and a point cloud, we compute superpoints, initial class-agnostic 3D instance masks, and per-view 2D masks; a 2D-guided refinement fuses multi-view segmentations into refined 3D masks. (2) For each 3D instance, we select top-$k$ views, encode them with an image–text encoder to obtain a multi-view image descriptor, and in parallel encode all candidate text labels; cosine similarities between image and text embeddings provide per-label scores. (3) {Selective VLM reasoning:} a temperature-scaled softmax over the cosine scores yields a routing score; high-confidence cases take the encoder top-1, while low-confidence cases pass the top-$K$ labels and the original images with bounding boxes to a reasoning VLM, which selects the final label using full-image context.}
\vspace{-8mm}
\label{method}
\end{center}
\end{figure*}

We address these limitations with Scene-Q, a confidence-aware, coarse-to-fine framework for interactive 3D scene querying. Scene-Q operates on class-agnostic 3D instance proposals~\cite{schult2023mask3d, kolodiazhnyi2024oneformer3d} and performs lightweight multi-view refinement to obtain robust instance representations. Each refined instance is encoded with multi-view image–text embeddings and passed through a confidence-aware routing module. We apply temperature-scaled softmax to the image–text embedding similarities to obtain normalized scores for routing decisions. High-confidence cases are resolved directly by the encoder via a fast path. When confidence in encoder predictions is low, the system optionally consults a reasoning Vison Language Model(VLM)~\cite{bai2025qwen2, liu2023visual, alayrac2022flamingo} for additional context-aware disambiguation. This step improves robustness in ambiguous cases: image–text encoders are retrieval-centric and weak at context-sensitive, compositional reasoning, whereas VLMs can leverage global and relational cues. The VLM operates on the original multi-view images with instance bounding boxes, enabling it to exploit scene context—for example, kitchen surroundings that distinguish \emph{dining table} from \emph{table}—and to disambiguate fine-grained classes, attributes, and states. The same routed reasoning applies symmetrically to both \emph{instance$\rightarrow$texts} and \emph{text$\rightarrow$instances} queries, as illustrated in Fig.~\ref{system_proposal}.

To evaluate the effectiveness of our method, we validate Scene-Q on ScanNet200, measuring open-vocabulary 3D instance segmentation and text-guided instance retrieval against encoder-only baselines using image–text encoders. Additionally, to demonstrate real-world applicability, we deploy Scene-Q in newly captured novel indoor environments, confirming generalization across diverse scenes and support for natural, open-ended interactions.

In summary, the contributions of our work are as follows:

\begin{itemize}
    \item A confidence-aware routing mechanism that applies temperature scaling to estimate confidence from image–text encoder similarities, and selectively invokes a reasoning VLM only for ambiguous instances.
    \item A context-aware multi-view VLM reasoning strategy that operates on full images with instance bounding boxes to leverage global scene context for fine-grained labeling and descriptive retrieval.
    \item Empirical gains over encoder-only baselines on standard open-vocabulary 3D benchmarks and real-world indoor captures, particularly on ambiguous categories and crowded scenes, highlighting the practical benefits of selective multimodal reasoning.
\end{itemize}

\section{Related Work}

\subsection{Image–Text Encoders}
Contrastively trained image–text encoders learn aligned embeddings from large image–caption corpora and are widely used for open-vocabulary recognition via cosine similarity~\cite{radford2021learningCLIP, sun2023eva, yu2022coca, zhai2023sigmoid, tschannen2025siglip}. These models excel at zero-shot retrieval but typically emit a single global embedding per image. To obtain instance-level descriptors, pipelines therefore crop or mask the object region in each view before encoding. This crop-based practice strips away global cues—room type, spatial layout, and inter-object relations—that are often decisive for choosing among near-synonymous labels. Moreover, similarity-based matching reflects retrieval strength rather than language reasoning about attributes, states, or relations, which yields tightly clustered scores and unstable top-1 selections when scenes contain many visually similar instances.

\subsection{Open-Vocabulary 3D Instance Segmentation}
Open-vocabulary 3D instance segmentation (OV3DIS) extends image–text encoders to 3D maps by aggregating per-instance features from multi-view observations and matching them to text labels~\cite{yang2023sam3d, peng2023openscene, kim2024ov, nguyen2024open3dis, yan2024maskclustering, huang2024openins3d, xu2025sampro3d, yin2024sai3d, lu2023ovir, boudjoghra2024openyolo3d}. Benchmarks such as ScanNet200 are evaluated by comparing hundreds of labels against each 3D instance via cosine similarity. In practice, near-synonymous categories lead to clustered scores, and many similar objects within a scene further confound top-1 matching. To improve geometry, many methods rely on class-agnostic 3D instance proposal networks such as Mask3D~\cite{schult2023mask3d} and OneFormer3D~\cite{kolodiazhnyi2024oneformer3d}, and then refine masks using foundation 2D segmentation fused across views~\cite{qi2022high, kirillov2023segment}. Yet the final semantic decision is often left to encoder-only matching, which remains brittle under ambiguity. Scene-Q addresses this gap by retaining efficient encoder decisions for high-confidence cases and escalating ambiguous ones to a reasoning model that operates with full-image context.

\subsection{Reasoning Vision–Language Models}
VLMs pair visual encoders with large language models (LLMs)~\cite{zhang2022opt, chung2024scaling, touvron2023llama, bai2023qwen}, effectively giving an LLM visual perception. Compared to image–text encoders that optimize for retrieval, VLMs inherit the LLM’s broad knowledge, instruction-following behavior, and emergent compositional skills, and then align these capabilities to vision~\cite{tsimpoukelli2021multimodal, wang2022ofa, lu2022unified, li2022blip, li2023blip, liu2023visual, alayrac2022flamingo, bai2025qwen2}. When applied to full images with localized regions, they can reason using global and relational cues to disambiguate fine-grained categories (e.g., \emph{table} vs. \emph{dining table}), recognize attributes or states, and follow natural-language instructions grounded in the scene. Region-aware VLMs further supervise grounding and spatial reasoning—producing pixel-grounded responses, conditioning on region proposals, or distilling spatial cognition from 3D scene graphs—thereby strengthening fine-grained understanding and multi-object reasoning~\cite{rasheed2024glamm, guo2024regiongpt, cheng2024spatialrgpt}. However, invoking a VLM on every query is unnecessary and computationally heavy, motivating selective designs that combine image–text encoders for easy, high-confidence cases with confidence-aware routing of ambiguous cases to a VLM that reasons over the original multi-view context.

\section{METHODS}
The overall pipeline is presented in Fig.~\ref{method}. Given registered RGB-D images and a reconstructed 3D point cloud, Scene-Q constructs a queryable 3D instance map. The system has two components: (i) a 3D instance segmentation module that produces class-agnostic 3D instance masks and refines them using multi-view 2D foundation segmentation; and (ii) a confidence-aware routing module that builds multi-view image--text descriptors and selectively invokes a reasoning VLM to resolve low-confidence cases using the original multi-view images together with regional instance localization.

\subsection{3D Instance Segmentation via 2D-Guided Refinement}
We first obtain class-agnostic 3D instance masks $\{M_i\}$ from a pre-trained 3D instance segmentation model~\cite{kolodiazhnyi2024oneformer3d, schult2023mask3d}. For each instance, we select the top-$k$ views where it is most visible and run a 2D foundation segmenter~\cite{kirillov2023segment, qi2022high}. Using the resulting multi-view 2D segmentations, we refine the initial 3D masks by (i) consolidating point assignments at the superpoint level, (ii) detaching geometrically separated components with low multi-view 2D agreement, and (iii) iteratively merging adjacent instances with high agreement until convergence, producing the refined set $\mathcal{X}$.

\textbf{Superpoint consolidation.}
To improve coherence and efficiency, we operate on superpoints (spatially adjacent points with similar geometry/color). Let $\mathcal{P}=\{p_n\}_{n=1}^{N}$ be the reconstructed point cloud and $\mathcal{S}=\{S_\ell\}_{\ell=1}^{N_S}$ a superpoint partition. We lift point-level instance assignments to superpoints by majority vote, yielding a compact representation that reduces spurious isolated points and accelerates subsequent refinement.

\textbf{Multi-view 2D agreement.}
We define a single multi-view consistency score used for both detachment and merging. For any two 3D regions $a$ and $b$ (superpoints, clusters, or instances), let $\mathcal{V}_{ab}$ be the intersection of their selected top-$k$ view sets, restricted to views where both are visible. In each view $v\in\mathcal{V}_{ab}$, we project points from $a$ and $b$ into the image and assign each region the dominant 2D segment ID by majority vote over its projected pixels, denoted $L_a(I_v)$ and $L_b(I_v)$. The agreement is
\vspace{-1mm}
\begin{equation}
\label{eq:agree}
\mathrm{Agree}(a,b)=\frac{1}{|\mathcal{V}_{ab}|}\sum_{v\in\mathcal{V}_{ab}} \mathbb{1}\!\left[L_a(I_v)=L_b(I_v)\right],
\end{equation}
and we set $\mathrm{Agree}(a,b)=0$ when $|\mathcal{V}_{ab}|=0$.

\textbf{Geometric split + 2D-verified detachment.}
Within each initial 3D instance $M$, we propose spatial components $\{c\}$ via a lightweight geometric clustering step~\cite{schubert2017dbscan} and denote the largest component as the main component $m$. To avoid over-splitting, we detach a candidate component $c\neq m$ into a new instance only when it is geometrically separated \emph{and} weakly supported by multi-view 2D agreement:
\vspace{-1mm}
\begin{equation}
\label{eq:detach_rule}
\text{detach } c \text{ if } \mathrm{Agree}(c,m) < \tau_{\text{detach}}.
\end{equation}

\textbf{2D-Driven merging of neighboring instances.}
After detachment, we form candidate neighboring instance pairs using 3D contact/adjacency between superpoints and merge adjacent instances that are highly consistent in multi-view 2D labels:
\vspace{-1mm}
\begin{equation}
\label{eq:merge_rule}
\text{merge } x_i, x_j \text{ if } \mathrm{Agree}(x_i,x_j) > \tau_{\text{merge}}.
\end{equation}
We apply this rule iteratively until no further merges occur, yielding the refined instance set $\mathcal{X}$.

\subsection{Selective VLM Reasoning}
Given a refined 3D instance $x$ and a label set $\mathcal{Y}$, Scene-Q predicts a semantic label via a two-stage, confidence-aware router. We first score all labels using a lightweight image--text encoder. If the encoder is confident, we directly return its top-1 prediction (fast path). Otherwise, we rerank a small set of top candidates using a reasoning VLM that observes the original multi-view images together with 2D bounding boxes localizing the instance. This design keeps inference efficient on easy cases while reserving expensive VLM reasoning for ambiguous instances.

\textbf{Multi-view instance descriptor.}
For each instance $x$, we select the top-$k$ views $\mathcal{V}_x$ where $x$ is most visible. In each selected view, a foundation segmenter mask~\cite{kirillov2023segment} is expanded by a fixed ratio across $L$ levels to form multi-level bounding boxes. We crop each box and encode all crops with an image--text encoder, then aggregate them into a single instance embedding:
\begin{equation}
\label{eq:descriptor}
f_{\mathrm{img}}(x)=\mathrm{Normalize}\!\left(\frac{1}{|\mathcal{C}_x|}\sum_{c\in\mathcal{C}_x}\phi(c)\right),
\vspace{-1mm}
\end{equation}
where $\mathcal{C}_x$ is the set of all crops for $x$ (over levels and views), $\phi(\cdot)$ is the encoder’s image embedding, and $\mathrm{Normalize}(\cdot)$ denotes $\ell_2$ normalization.

\textbf{Confidence estimation and selective routing.}
We embed each label $y_j\in\mathcal{Y}$ with the encoder’s text tower to obtain $f_{\mathrm{text}}(y_j)$, and compute cosine similarities
\begin{equation}
\label{eq:sj}
s_j=\cos\!\big(f_{\mathrm{img}}(x), f_{\mathrm{text}}(y_j)\big).
\end{equation}
We convert similarities into a temperature-scaled distribution via a softmax with temperature $T$:
\begin{equation}
\label{eq:softmax}
p_j=\frac{\exp(s_j/T)}{\sum_{t}\exp(s_t/T)}.
\end{equation}
From $p=\{p_j\}$ we compute three confidence statistics: the maximum probability $p_{\max}$, the top-1/top-2 margin $m=p_{(1)}-p_{(2)}$ (where $p_{(1)}$ and $p_{(2)}$ are the largest and second-largest probabilities), and the normalized entropy $H_{\mathrm{norm}}=H(p)/\log C$ with $C=|\mathcal{Y}|$:
\begin{equation}\label{eq:stats}
p_{\max}=\max_{j}p_j,\;\; m=p_{(1)}-p_{(2)},\;\; H_{\mathrm{norm}}=\frac{H(p)}{\log C},
\end{equation}
where $H(p)=-\sum_{j=1}^{C} p_j\log p_j$.
We take the encoder fast path when the prediction is sharp and well-separated according to the following routing criterion:
\begin{equation}\label{eq:gate}
p_{\max}\ge \delta_p \;\land\; m\ge \delta_m \;\land\; \big(H_{\mathrm{norm}}\le \delta_H \;\lor\; m\ge \delta_{em}\big).
\end{equation}
We set $\delta_{em}>\delta_m$ so that an exceptionally large margin can override entropy. If the routing criterion is not satisfied, we form a candidate set $\mathcal{Y}_K$ from the top-$K$ labels by $s_j$ and defer to VLM reasoning.

\textbf{VLM reasoning with full-image context.}
For low-confidence instances, we query a reasoning VLM with the {original} multi-view images $\mathcal{V}_x$. In each view, we provide a single 2D bounding box that localizes $x$ and pose a multiple-choice prompt over $\mathcal{Y}_K$, instructing the model to select exactly one label for the boxed region. We aggregate per-view predictions by majority vote to obtain $y^\star$; ties are resolved by falling back to the encoder’s top-1 label within $\mathcal{Y}_K$.

\begin{table*}[t!]
\centering
\vspace{2mm}
\caption{3D instance segmentation results on ScanNet200~\cite{rozenberszki2022language}.}
\label{tab:3d-instance-segmentation}

\begin{tabular}{@{}lccccccccc@{}} 
\toprule
\textbf{Method} & \textbf{Query Type} & \textbf{3D Network} & \textbf{2D--3D Refinement} & \textbf{mAP} & \textbf{AP$_{50}$} & \textbf{AP$_{25}$} & \textbf{head (AP)} & \textbf{common (AP)} & \textbf{tail (AP)} \\
\midrule
\rowcolor{black!6}
\cv{ISBNet~\cite{ngo2023isbnet}} & \cv{} & \cv{--} & \cv{--} & \cv{24.5} & \cv{32.7} & \cv{37.6} & \cv{38.6} & \cv{20.5} & \cv{12.5} \\
\rowcolor{black!6}
\cv{Mask3D~\cite{schult2023mask3d}} & \cv{\textbf{Closed-vocab}} & \cv{--} & \cv{--} & \cv{26.9} & \cv{36.2} & \cv{41.4} & \cv{39.8} & \cv{21.7} & \cv{17.9} \\
\rowcolor{black!6}
\cv{Oneformer3D~\cite{kolodiazhnyi2024oneformer3d}} & \cv{} & \cv{--} & \cv{--} & \cv{30.6} & \cv{40.8} & \cv{45.4} & \cv{44.7} & \cv{24.6} & \cv{21.0} \\
\midrule
OpenScene~\cite{peng2023openscene} & \multirow{6}{*}{Open-vocab} & \xmark & \xmark & 2.8 & 7.8 & 18.6 & 2.7 & 3.1 & 2.6 \\
SAM3D~\cite{yang2023sam3d} &  & \xmark & \cmark & 8.4 & 13.1 & 18.7 & 9.3 & 7.0 & 9.1 \\
OV-MAP~\cite{kim2024ov} &  & \xmark & \cmark & 11.9 & 17.4 & 23.2 & 12.5 & 10.5 & 12.7 \\
SAI3D~\cite{yin2024sai3d} &  & \xmark & \cmark & 12.7 & 18.8 & 24.1 & 12.1 & 10.4 & 16.2 \\
OVIR-3D~\cite{lu2023ovir} &  & \xmark & \cmark & 13.0 & 24.9 & 32.3 & 14.4 & 12.7 & 11.7 \\
Open3DIS~\cite{nguyen2024open3dis} &  & \xmark & \cmark & 18.2 & 26.1 & 31.4 & 18.9 & 16.5 & 19.2 \\

\midrule
OpenIns3D~\cite{huang2024openins3d} & \multirow{6}{*}{Open-vocab} & \cmark & \cmark & 8.8 & 10.3 & 14.4 & 16.0 & 6.5 & 4.2 \\
OpenScene~\cite{peng2023openscene} &  & \cmark & \xmark & 11.7 & 15.2 & 17.8 & 13.4 & 11.6 & 9.9 \\
OpenMask3D~\cite{takmaz2023openmask3d} &  & \cmark & \xmark & 15.4 & 19.9 & 23.1 & 17.1 & 14.1 & 14.9 \\
Open3DIS~\cite{nguyen2024open3dis} &  & \cmark & \cmark & 23.7 & 29.4 & 32.8 & 27.8 & 21.2 & 21.8 \\
OpenYOLO3D~\cite{boudjoghra2024openyolo3d} &  & \cmark & \xmark & 24.5 & 31.7 & 36.2 &  27.8 & 24.3 & 21.6 \\
\textbf{Ours} &  & \cmark & \cmark & \textbf{25.2} & \textbf{33.7} & \textbf{41.6} & 25.9 & 24.2 & 25.7 \\
\bottomrule
\end{tabular}

\vspace{2pt}
\noindent\hspace*{-4.5em}
\begin{minipage}{0.9\textwidth}
\footnotesize
\textbf{Best} results per column are in \textbf{bold}. Shaded rows (Closed-vocab) are not directly comparable to Open-vocab.
\end{minipage}
\vspace{-6mm}
\end{table*}

\section{Experiments}\label{experiments}
We evaluate Scene-Q with quantitative comparisons, ablations, and qualitative analyses on two complementary directions of language grounding: open-vocabulary 3D instance segmentation (\emph{instance$\rightarrow$texts}) and natural-language 3D instance retrieval (\emph{text$\rightarrow$instances}). 
In Sec.~\ref{sec:exp-ov3dis}, we compare against open-vocabulary 3D instance segmentation baselines and present ablations. 
In Sec.~\ref{sec:exp-qual}, we show qualitative results highlighting class-agnostic 3D instance refinement and selective VLM reasoning.

\textbf{Datasets.}
For \emph{instance$\rightarrow$texts} evaluation, we use ScanNet200 and report results on the validation split (312 scenes, 200 categories). We also analyze long-tail behavior using the standard head/common/tail grouping (66/68/66).
For \emph{text$\rightarrow$instances} evaluation, we curate a real-world retrieval set by annotating human-authored natural-language queries with a single target instance on reconstructed novel maps. The maps are captured with Azure Kinect and reconstructed with RTAB-Map~\cite{labbe2019rtab}. We generate initial instance candidates using OV-Map~\cite{kim2024ov}, and then manually refine/verify instance masks and query--target pairs to enable unambiguous \emph{top-1} retrieval evaluation. The dataset contains 11 scenes and 1968 queries in total (Category: 547, Attribute: 601, Spatial: 381, Affordance: 439).

\textbf{Metrics.}
For ScanNet200 open-vocabulary 3D instance segmentation, we follow the standard ScanNet200 evaluation protocol and report AP@50 and AP@25 (mask IoU thresholds of 50\% and 25\%, respectively), as well as mAP averaged over IoU thresholds from 0.50 to 0.95 in increments of 0.05. For natural-language 3D instance retrieval, we report Hit@1 (top-1 accuracy): given a query, a method selects a single 3D instance from the scene and is counted as correct if it matches the annotated target. We evaluate retrieval in two settings: (i) \emph{with GT instance masks}, where candidates are annotated instance masks and correctness is an exact instance match, and (ii) \emph{without GT instance masks}, where candidates are each method's predicted instance masks and a prediction is counted as correct if the selected instance overlaps the annotated target with IoU $\geq 0.5$.

\textbf{Implementation details.} We process posed RGB-D sequences, subsampling every 10th frame and selecting the top-5 views per 3D instance (by visibility). Initial class-agnostic 3D instance masks are obtained from pretrained 3D instance segmentation models~\cite{kolodiazhnyi2024oneformer3d, schult2023mask3d}, and per-view 2D masks are produced by a foundation 2D segmenter~\cite{kirillov2023segment, qi2022high}. We refine 3D instances with a geometric detachment step based on DBSCAN~\cite{schubert2017dbscan}. With 2D masks precomputed for the selected views, refinement takes about 2--4 minutes per scene. For selective VLM reasoning, we use image--text encoders~\cite{radford2021learningCLIP, tschannen2025siglip} for initial scoring and Qwen2.5-VL~\cite{bai2025qwen2} as the reasoner. 

We use a held-out tuning subset (10\% of the training split) for temperature $T$ and routing threshold tuning. The $T$ is fit by minimizing negative log-likelihood, and the routing thresholds ($\delta_p\!=\!0.50$, $\delta_m\!=\!0.15$, $\delta_H\!=\!0.35$, $\delta_{em}\!=\!0.20$) are selected to maximize the F1 score of the routing decision, balancing efficiency and accuracy. Sensitivity analysis on this set confirms that performance is stable, with mAP fluctuations of less than $\pm$0.6 for threshold variations of $\pm$10\%, indicating that the method is robust to hyperparameter choice. We set $K\!=\!10$ for top-$K$ reasoning. Semantic annotation (offline indexing) takes about 3--5 minutes per scene on a single NVIDIA RTX~8000 GPU. Crucially, this is a one-time indexing cost. During online interaction, high-confidence queries routed through the encoder fast path take roughly 50ms, while low-confidence queries that invoke VLM reasoning take $\sim$2s. Let $r$ denote the fraction of queries escalated to the VLM. The expected per-query latency is then $(1-r)\cdot0.05 + r\cdot2.0$ seconds, corresponding to about 0.81–1.26s/query at the observed routing rate of 39--62\%. This selective latency remains acceptable for mobile robotics, where physical navigation time typically dominates computational overhead.

\subsection{Quantitative Results}\label{sec:exp-ov3dis}

\textbf{ScanNet200.}
For the \emph{instance$\rightarrow$texts} direction—i.e., open-vocabulary label assignment for 3D instances—we evaluate open-vocabulary 3D instance segmentation on the ScanNet200 validation split (Table~\ref{tab:3d-instance-segmentation}). 
Scene-Q yields its largest gains under relaxed IoU, achieving AP@25 of 41.6 and improving by +5.4 over OpenYOLO3D~\cite{boudjoghra2024openyolo3d} (36.2). This suggests improved robustness to noisy or imperfect masks commonly encountered in practice.
Overall, Scene-Q achieves mAP of 25.2 and AP@50 of 33.7, corresponding to gains of +0.7 mAP and +2.0 AP@50. We observe that the performance gap widens as the IoU threshold is relaxed, with the largest margin at AP@25. We attribute this trend to the quality of visual evidence at different thresholds: high-IoU settings typically correspond to clean, well-isolated object observations, whereas AP@25 better reflects harsher real-world cases with loose or noisy masks. In these low-IoU regimes, prior CLIP~\cite{radford2021learningCLIP}-based methods often struggle because they rely on cropped object regions, which can contain misleading context or insufficient discriminative features when masks are imperfect. In contrast, Scene-Q does not rely solely on crops: the VLM analyzes the full image with bounding-box overlays, leveraging global context to identify objects correctly even when pixel-level isolation is imperfect.

Crucially, this access to global context serves a dual purpose: it compensates for poor segmentation and simultaneously provides the necessary cues for fine-grained distinction. By leveraging this context to resolve hard semantic cases—such as distinguishing near-synonyms or resolving context-dependent labels—Scene-Q achieves robust retrieval (AP@25) crucial for robotic interaction, where identifying the correct object instance is the primary failure mode of existing systems.

\begin{figure*} [t!]
\begin{center}
\vspace{2mm}
\includegraphics[width=0.89\textwidth]{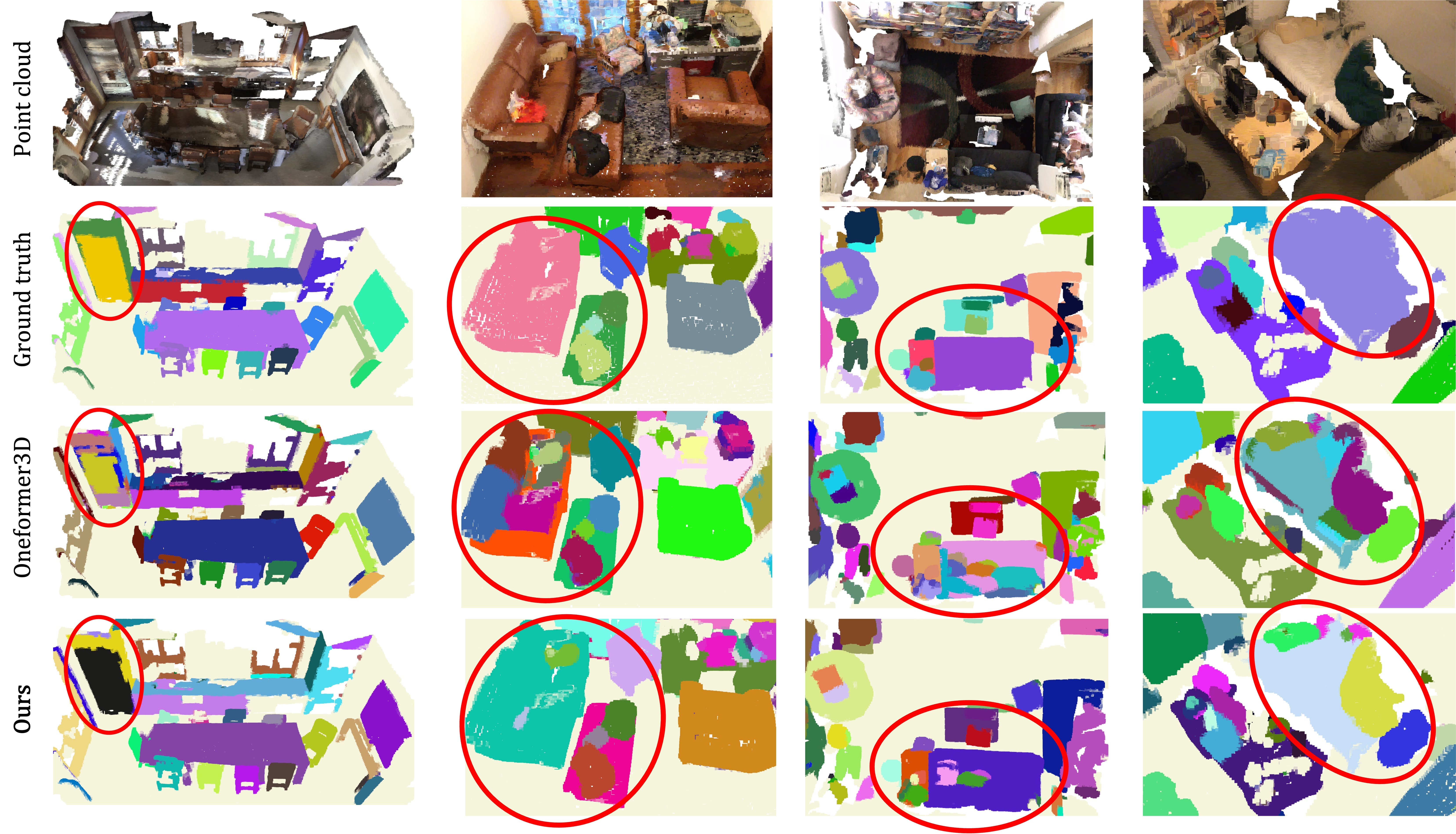}
\vspace{-2mm}
\caption{\textbf{Qualitative Segmentation Results.} The images illustrate ScanNet200 input scenes, ground truth (GT) annotations, and our model's class-agnostic 3D instance segmentation results, refined through our 2D-guided refinement process.}
\label{seg_qual_res}
\vspace{-7mm}
\end{center}
\end{figure*}

Furthermore, we observe that standard benchmarks penalize ``better-than-ground truth (GT)" segmentations. Qualitative inspection (Fig.~\ref{seg_qual_res}) reveals that GT annotations often lump distinct objects together (e.g., a sofa and the items resting on it), whereas our method correctly segments them as separate instances. Because the benchmark penalizes this over-segmentation against the coarse GT, our quantitative gains are artificially suppressed. This suggests that the modest +0.7 mAP gain understates the potential semantic advantage of our method, which is robust not only to geometric noise but also to annotation ambiguity.

Because open-vocabulary methods are not trained with per-class supervision, head/common/tail splits are less diagnostic for cross-method comparisons among open-vocabulary models. Instead, we contrast with closed-vocabulary references that use full supervision and typically show strong head bias. Mask3D reports head/common/tail AP of 39.8/21.7/17.9 (gap: 21.9), whereas OneFormer3D reports 44.7/24.6/21.0 (gap: 23.7). In contrast, our method maintains a balanced performance profile (head/tail AP: 25.9/25.7; gap: 0.2), indicating that our reliance on VLM reasoning effectively transfers knowledge to tail categories without the need for extensive training data.

Closed-vocabulary models on ScanNet200 are not directly comparable because they rely on full supervision for a fixed label set, but we include them as references. We exceed ISBNet (mAP 24.5, AP@50 32.7, AP@25 37.6) by +0.7 mAP, +1.0 AP@50, and +4.0 AP@25. Relative to Mask3D (mAP 26.9, AP@50 36.2, AP@25 41.4), our mAP is 1.7 points lower and AP@50 is 2.5 lower, while AP@25 is slightly higher (+0.2). This suggests that our open-vocabulary masks remain competitive despite the harder experimental setting.

\textbf{Ablation study: Decoupling semantics from geometry.}
Standard mAP metrics (Table~\ref{tab:3d-instance-segmentation}) conflate geometric instance quality with semantic classification accuracy, which can obscure improvements in semantic understanding. To isolate our semantic contribution, we evaluate on ScanNet200 using oracle (GT) instance masks (Table~\ref{ablation_exp}).

When geometric noise is removed, the benefit of selective VLM reasoning becomes clear. Relative to an encoder-only pipeline (\textsc{None}), adding our reasoning stage (\textsc{Selective}) yields large gains: with a CLIP backbone, mAP increases from 32.9 to 43.6 (+32.5\% relative), and with SigLIP~\cite{tschannen2025siglip} it improves from 48.4 to 51.9 (+7.2\%). This contrast is also evident against strong baselines: in the standard end-to-end setting with predicted masks (Table~\ref{tab:3d-instance-segmentation}), Scene-Q improves by only +0.7 mAP over the strongest baseline, whereas under oracle masks (Table~\ref{ablation_exp}) it reaches 51.9 mAP compared to 39.6 mAP for OpenYOLO3D (+12.3 mAP). This indicates that geometric proposal errors dominate the end-to-end benchmark and can substantially suppress the observable impact of improved semantic reasoning.

Interestingly, invoking the VLM for all instances (\textsc{Always}) is slightly worse than selective routing (42.8 vs.\ 43.6 with CLIP; 50.8 vs.\ 51.9 with SigLIP). This suggests that VLM reasoning is most valuable for ambiguous instances: when the encoder is already confident from local evidence, forcing global-context reasoning can occasionally distract the model and flip an otherwise correct prediction. In practice, \textsc{Selective} triggers the VLM only for the uncertain subset (39--62\% of instances per scene), yielding a better accuracy--efficiency trade-off and higher AP improvement per VLM call.

Overall, this ablation exposes the limitation of standard image--text encoders: while they propose a coarse candidate set, they often fail at fine-grained disambiguation without global context. Our selective VLM reasoning resolves these cases. Even without dataset-specific training, our oracle performance (51.9 mAP with SigLIP under \textsc{Selective}) substantially exceeds fully supervised closed-vocabulary references such as Mask3D (35.5 mAP), highlighting that once geometry is fixed, semantic reasoning becomes the primary driver of performance.

\begin{table}[t!]
\centering
\vspace{2mm}
\caption{Ablation of Selective VLM reasoning on ScanNet200 3D instance segmentation with oracle masks.}
\label{ablation_exp}
\setlength{\tabcolsep}{4pt}
\begin{tabular}{@{}l c c c c @{}}
\toprule
\textbf{Method} &
\textbf{\shortstack[c]{Semantic\\2D Model}} &
\textbf{\shortstack[c]{VLM\\Reasoning}} &
\textbf{mAP} &
\textbf{\shortstack[c]{\(\Delta\)mAP (\%) vs.\\\textsc{None}}} \\
\midrule

\multicolumn{5}{@{}l@{}}{\textit{Closed-vocab}} \\
Mask3D~\cite{schult2023mask3d} & -- & -- & 35.5 & -- \\
\midrule

\multicolumn{5}{@{}l@{}}{\textit{Open-vocab}} \\
OpenScene~\cite{peng2023openscene} & \cite{ghiasi2022scaling} & -- & 22.9 & -- \\
OpenMask3D~\cite{takmaz2023openmask3d} & \cite{radford2021learningCLIP} & -- & 29.1 & -- \\
Open3DIS~\cite{nguyen2024open3dis} & \cite{radford2021learningCLIP, ren2024grounded} & -- & 30.9 & -- \\
OV-Map~\cite{kim2024ov} & \cite{radford2021learningCLIP} & -- & 31.2 & -- \\
OpenYOLO3D~\cite{boudjoghra2024openyolo3d} & \cite{cheng2024yoloworld} & -- & 39.6 & -- \\
\midrule

\multirow{6}{*}{\textbf{Scene-Q (Ours)}} &
\multirow{3}{*}{\cite{radford2021learningCLIP}} &
\textsc{None}      & 32.9 & -- \\
& & \textsc{Selective} & \textbf{43.6} & \textbf{+32.5\%} \\ 
& & \textsc{Always}    & 42.8 & +30.1\% \\ 
\cmidrule(l){2-5}
\multirow{6}{*}{} &
\multirow{3}{*}{\cite{tschannen2025siglip}} &
\textsc{None}      & 48.4 & -- \\
& & \textsc{Selective} & \textbf{51.9} & \textbf{+7.2\%} \\  
& & \textsc{Always}    & 50.8 & +5.0\% \\  
\bottomrule
\end{tabular}
\vspace{-4mm}
\end{table}

\textbf{Real-world.}
For the \emph{text$\rightarrow$instances} direction—i.e., natural-language 3D instance retrieval—we evaluate on real-world reconstructed novel maps with free-form, human-authored queries. 
All methods use 3D instance proposal/mask models applied off-the-shelf from ScanNet200-trained checkpoints, without finetuning or parameter retuning on our maps, providing a direct test of robustness under distribution shift.
Queries are grouped into four buckets: Category (exact class names), Attribute/appearance (e.g., color/material), Spatial relation, and Affordance/high-level intent.
We report Hit@1 under two settings (Table~\ref{tab:realworld-retrieval}): (i) \emph{with GT instance masks}, where candidates are annotated instance masks, and (ii) \emph{without GT instance masks}, where candidates are the instance masks predicted by each method.

Across both settings, Scene-Q consistently outperforms prior methods, with particularly large gains on spatial and affordance queries. 
In the GT-mask setting, baselines perform reasonably on category-name queries but degrade sharply on spatial and affordance, where crop-based matching often fails to leverage room context and inter-object relations. 
By selectively invoking VLM reasoning on low-confidence cases while providing full-image context with box localization, Scene-Q substantially improves performance on these compositional queries.

In the predicted-mask setting, overall accuracy drops for all methods due to errors in out-of-distribution 3D instance proposals and masks, yet Scene-Q remains strong and preserves a large margin over baselines. 
This indicates that the proposed method is beneficial not only when instance masks are clean, but also under realistic segmentation noise.

\begin{table}[t]
\centering
\vspace{2mm}
\caption{Natural-language instance retrieval on real-world reconstructed novel maps.}
\label{tab:realworld-retrieval}
\setlength{\tabcolsep}{5.5pt}
\begin{tabular}{@{}lcccc@{}}
\toprule
\textbf{Method} &
\textbf{Category} &
\textbf{Attribute} &
\textbf{Spatial} &
\textbf{Affordance} \\
\midrule
\multicolumn{5}{@{}l}{\textit{With GT instance masks}} \\
OpenMask3D~\cite{takmaz2023openmask3d} & 61.4 & 51.6 & 15.2 & 5.7 \\
Open3DIS~\cite{nguyen2024open3dis}     & 68.6 & 52.2 & 16.8 & 5.9 \\
OpenYOLO3D~\cite{boudjoghra2024openyolo3d} & 71.7 & 31.4 & 2.1 & 1.4 \\
\textbf{Scene-Q (Ours)} & \textbf{75.9} & \textbf{69.2} & \textbf{61.4} & \textbf{41.2} \\
\midrule
\multicolumn{5}{@{}l}{\textit{Without GT instance masks (predicted masks)}} \\
OpenMask3D~\cite{takmaz2023openmask3d} & 19.2 & 16.6 & 3.7 & 1.1 \\
Open3DIS~\cite{nguyen2024open3dis}     & 40.8 & 29.3 & 10.2 & 3.4 \\
OpenYOLO3D~\cite{boudjoghra2024openyolo3d} & 53.2 & 21.5 & 1.6 & 1.1 \\
\textbf{Scene-Q (Ours)} & \textbf{67.6} & \textbf{61.6} & \textbf{54.6} & \textbf{36.7} \\
\bottomrule
\end{tabular}
\vspace{-6mm}
\end{table}

\subsection{Qualitative Results}\label{sec:exp-qual}
\textbf{ScanNet200.} Fig.~\ref{seg_qual_res} presents qualitative examples demonstrating the effectiveness of our approach in 3D instance segmentation. Scene-Q generates high-fidelity 3D instance masks by refining noisy initial predictions through our 2D-guided process. The results highlight the successful consolidation of over-segmented instances; for example, a sofa initially fragmented into multiple disconnected parts is correctly merged into a single coherent instance. Additionally, our boundary refinement sharpens object contours, leading to more precise segmentation compared to the coarse initial proposals.

Notably, as hypothesized in Sec.~\ref{sec:exp-ov3dis}, our method frequently surpasses the granularity of human-annotated GT. Visual inspection reveals that GT annotations often lump distinct objects together—such as grouping items placed on a bed or sofa into a single ``furniture" entity. In contrast, Scene-Q successfully separates these into distinct instances, yielding a more accurate semantic representation of the scene. This ability to resolve fine-grained distinct objects, even where training labels are coarse or missing, supports reliable open-vocabulary querying and highlights the potential of our approach for deployment in novel environments where high-quality 3D annotations are unavailable.

\begin{figure} [t!]
\begin{center}
\includegraphics[width=\columnwidth]{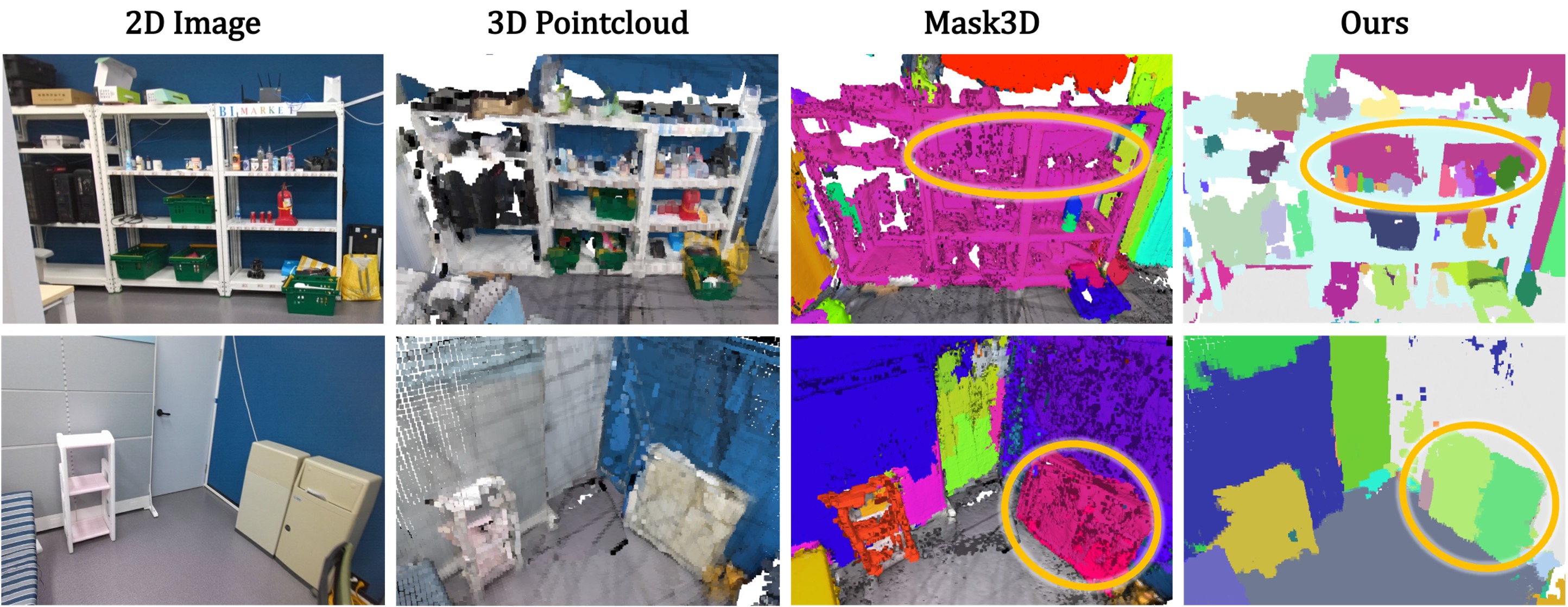}
\vspace{-6mm}
\caption{\textbf{Generalization on a Real-World Novel Map.} Comparison of ScanNet200-trained 3D proposals and Scene-Q refinement on a previously unseen map captured in our lab, with no parameter retuning.
Under domain shift, Mask3D proposals often merge instances or miss small objects, whereas Scene-Q separates fine-grained instances and preserves object boundaries, providing more reliable instance geometry for open-vocabulary querying.}
\label{real_ex}
\end{center}
\vspace{-9mm}
\end{figure}

\begin{figure} [t!]
\begin{center}
\includegraphics[width=\columnwidth]{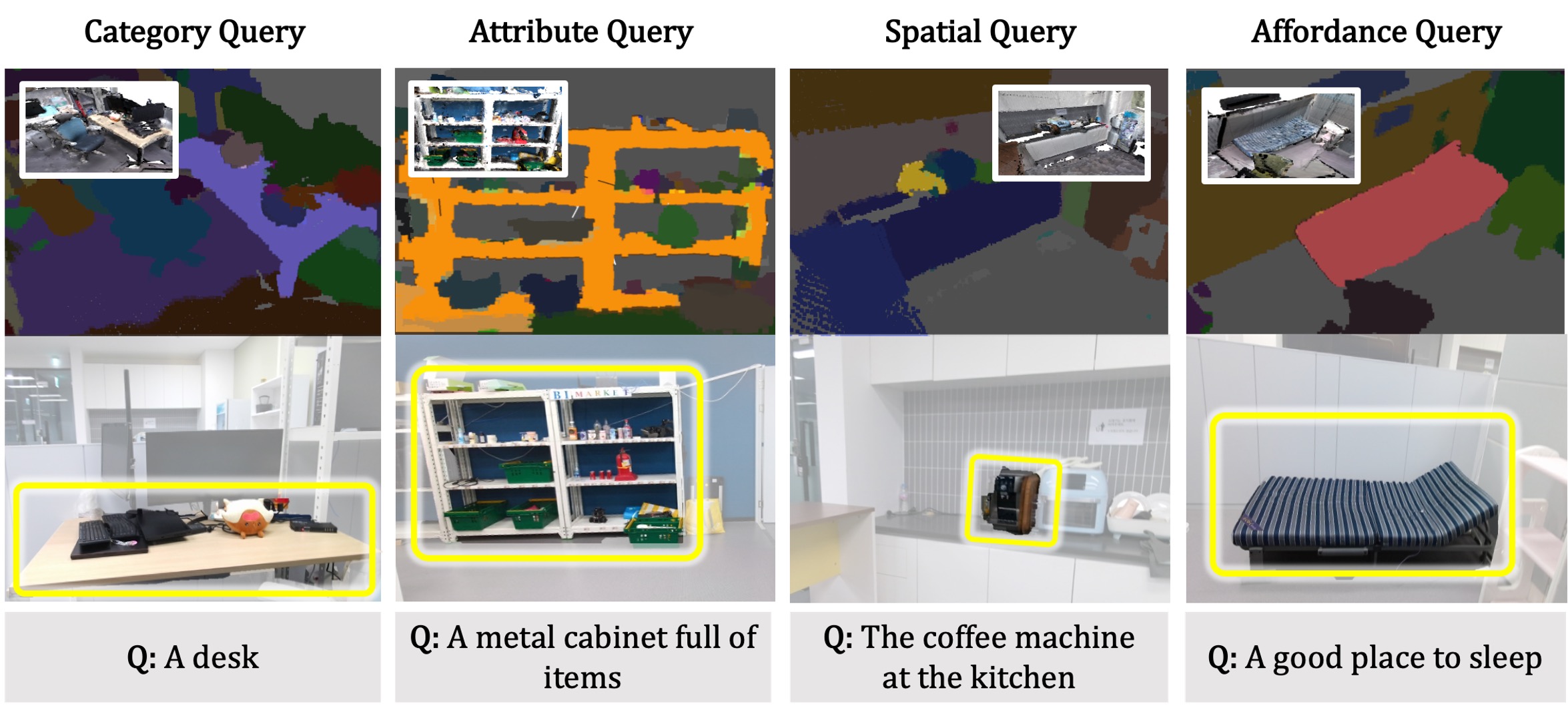} 
\vspace{-6mm}
\caption{\textbf{Qualitative Query Results on a Real-World Novel Maps.} Examples of \emph{text$\rightarrow$instances} retrieval on reconstructed real-world maps. Columns represent the four query categories in our benchmark: Category, Attribute/appearance, Spatial relation, and Affordance/high-level intent. For each query, the top row shows the retrieved target's 3D instance segmentation, while the bottom row shows a corresponding RGB view with the target highlighted by a yellow box. The examples demonstrate retrieval from free-form human-authored queries beyond exact category names.}
\label{qual_real}
\vspace{-9mm}
\end{center}
\end{figure}

\textbf{Real-world.}
We further evaluate Scene-Q on a previously unseen 3D map captured in our lab environment to assess generalization beyond ScanNet200. We use the exact same settings as in ScanNet200---including temperature scaling and routing thresholds---with no retuning on the novel map. Fig.~\ref{real_ex} shows that our refinement maintains reliable instance geometry under domain shift, mitigating common errors of ScanNet200-trained supervised proposals such as merged or missing instances. This improved geometric consistency provides a stronger foundation for language grounding. Building on this, Fig.~\ref{qual_real} presents representative successes across the four query buckets: category-name queries are typically high-confidence and handled by the fast path, while ambiguous attribute queries trigger VLM refinement to resolve fine-grained appearance cues using full-image context and bounding-box localization. Spatial queries benefit from global layout and inter-object relations across views, and affordance queries demonstrate higher-level intent reasoning that maps requests to plausible object types and disambiguates among candidates using surrounding context. The supplementary video further provides an end-to-end demonstration of Scene-Q in a multi-room environment.

\section{Limitations}
\looseness=-2
While Scene-Q demonstrates strong performance on both benchmarks and real-world deployments, two limitations remain. 
First, the system assumes a static scene; dynamic changes during or after capture can produce stale or inconsistent instance maps without explicit temporal tracking and map updates. 
Second, while confidence-based routing reduces VLM calls, future work will explore adaptive, budget-aware policies and larger-scale accuracy--cost analysis.
 
\FloatBarrier

\bibliographystyle{ieeetr}
\bibliography{ref}

@article{lu2022unified,
  title={Unified-io: A unified model for vision, language, and multi-modal tasks},
  author={Lu, Jiasen and Clark, Christopher and Zellers, Rowan and Mottaghi, Roozbeh and Kembhavi, Aniruddha},
  journal={arXiv preprint arXiv:2206.08916},
  year={2022}
}

@article{li2023blip,
  title={Blip-2: Bootstrapping language-image pre-training with frozen image encoders and large language models},
  author={Li, Junnan and Li, Dongxu and Savarese, Silvio and Hoi, Steven},
  journal={arXiv preprint arXiv:2301.12597},
  year={2023}
}

@inproceedings{wang2022ofa,
  title={Ofa: Unifying architectures, tasks, and modalities through a simple sequence-to-sequence learning framework},
  author={Wang, Peng and Yang, An and Men, Rui and Lin, Junyang and Bai, Shuai and Li, Zhikang and Ma, Jianxin and Zhou, Chang and Zhou, Jingren and Yang, Hongxia},
  booktitle={International Conference on Machine Learning},
  pages={23318--23340},
  year={2022},
  organization={PMLR}
}

@inproceedings{ghiasi2022scaling,
  title={Scaling open-vocabulary image segmentation with image-level labels},
  author={Ghiasi, Golnaz and Gu, Xiuye and Cui, Yin and Lin, Tsung-Yi},
  booktitle={European Conference on Computer Vision},
  pages={540--557},
  year={2022},
  organization={Springer}
}

@inproceedings{peng2023openscene,
  title={Openscene: 3d scene understanding with open vocabularies},
  author={Peng, Songyou and Genova, Kyle and Jiang, Chiyu and Tagliasacchi, Andrea and Pollefeys, Marc and Funkhouser, Thomas and others},
  booktitle={Proceedings of the IEEE/CVF Conference on Computer Vision and Pattern Recognition},
  pages={815--824},
  year={2023}
}

@article{takmaz2023openmask3d,
  title={Openmask3d: Open-vocabulary 3d instance segmentation},
  author={Takmaz, Ay{\c{c}}a and Fedele, Elisabetta and Sumner, Robert W and Pollefeys, Marc and Tombari, Federico and Engelmann, Francis},
  journal={arXiv preprint arXiv:2306.13631},
  year={2023}
}

@inproceedings{dai2017scannet,
  title={Scannet: Richly-annotated 3d reconstructions of indoor scenes},
  author={Dai, Angela and Chang, Angel X and Savva, Manolis and Halber, Maciej and Funkhouser, Thomas and Nie{\ss}ner, Matthias},
  booktitle={Proceedings of the IEEE conference on computer vision and pattern recognition},
  pages={5828--5839},
  year={2017}
}

@inproceedings{schult2023mask3d,
  title={Mask3d: Mask transformer for 3d semantic instance segmentation},
  author={Schult, Jonas and Engelmann, Francis and Hermans, Alexander and Litany, Or and Tang, Siyu and Leibe, Bastian},
  booktitle={2023 IEEE International Conference on Robotics and Automation (ICRA)},
  pages={8216--8223},
  year={2023},
  organization={IEEE}
}

@inproceedings{kirillov2023segment,
  title={Segment anything},
  author={Kirillov, Alexander and Mintun, Eric and Ravi, Nikhila and Mao, Hanzi and Rolland, Chloe and Gustafson, Laura and Xiao, Tete and Whitehead, Spencer and Berg, Alexander C and Lo, Wan-Yen and others},
  booktitle={Proceedings of the IEEE/CVF International Conference on Computer Vision},
  pages={4015--4026},
  year={2023}
}

@article{qi2022high,
  title={High-quality entity segmentation},
  author={Qi, Lu and Kuen, Jason and Guo, Weidong and Shen, Tiancheng and Gu, Jiuxiang and Jia, Jiaya and Lin, Zhe and Yang, Ming-Hsuan},
  journal={arXiv preprint arXiv:2211.05776},
  year={2022}
}

@article{yang2023sam3d,
  title={Sam3d: Segment anything in 3d scenes},
  author={Yang, Yunhan and Wu, Xiaoyang and He, Tong and Zhao, Hengshuang and Liu, Xihui},
  journal={arXiv preprint arXiv:2306.03908},
  year={2023}
}

@inproceedings{radford2021learningCLIP,
  title={Learning transferable visual models from natural language supervision},
  author={Radford, Alec and Kim, Jong Wook and Hallacy, Chris and Ramesh, Aditya and Goh, Gabriel and Agarwal, Sandhini and Sastry, Girish and Askell, Amanda and Mishkin, Pamela and Clark, Jack and others},
  booktitle={International conference on machine learning},
  pages={8748--8763},
  year={2021},
  organization={PMLR}
}

@inproceedings{rozenberszki2022language,
    title={Language-Grounded Indoor 3D Semantic Segmentation in the Wild},
    author={Rozenberszki, David and Litany, Or and Dai, Angela},
    booktitle = {Proceedings of the European Conference on Computer Vision ({ECCV})},
    year={2022}
}

@article{schubert2017dbscan,
  title={DBSCAN revisited, revisited: why and how you should (still) use DBSCAN},
  author={Schubert, Erich and Sander, J{\"o}rg and Ester, Martin and Kriegel, Hans Peter and Xu, Xiaowei},
  journal={ACM Transactions on Database Systems (TODS)},
  volume={42},
  number={3},
  pages={1--21},
  year={2017},
  publisher={ACM New York, NY, USA}
}

@inproceedings{kolodiazhnyi2024oneformer3d,
  title={Oneformer3d: One transformer for unified point cloud segmentation},
  author={Kolodiazhnyi, Maxim and Vorontsova, Anna and Konushin, Anton and Rukhovich, Danila},
  booktitle={Proceedings of the IEEE/CVF Conference on Computer Vision and Pattern Recognition},
  pages={20943--20953},
  year={2024}
}

@article{chung2024scaling,
  title={Scaling instruction-finetuned language models},
  author={Chung, Hyung Won and Hou, Le and Longpre, Shayne and Zoph, Barret and Tay, Yi and Fedus, William and Li, Yunxuan and Wang, Xuezhi and Dehghani, Mostafa and Brahma, Siddhartha and others},
  journal={Journal of Machine Learning Research},
  volume={25},
  number={70},
  pages={1--53},
  year={2024}
}

@article{zhang2022opt,
  title={Opt: Open pre-trained transformer language models},
  author={Zhang, Susan and Roller, Stephen and Goyal, Naman and Artetxe, Mikel and Chen, Moya and Chen, Shuohui and Dewan, Christopher and Diab, Mona and Li, Xian and Lin, Xi Victoria and others},
  journal={arXiv preprint arXiv:2205.01068},
  year={2022}
}

@article{alayrac2022flamingo,
  title={Flamingo: a visual language model for few-shot learning},
  author={Alayrac, Jean-Baptiste and Donahue, Jeff and Luc, Pauline and Miech, Antoine and Barr, Iain and Hasson, Yana and Lenc, Karel and Mensch, Arthur and Millican, Katherine and Reynolds, Malcolm and others},
  journal={Advances in neural information processing systems},
  volume={35},
  pages={23716--23736},
  year={2022}
}

@inproceedings{li2022blip,
  title={Blip: Bootstrapping language-image pre-training for unified vision-language understanding and generation},
  author={Li, Junnan and Li, Dongxu and Xiong, Caiming and Hoi, Steven},
  booktitle={International conference on machine learning},
  pages={12888--12900},
  year={2022},
  organization={PMLR}
}

@article{tsimpoukelli2021multimodal,
  title={Multimodal few-shot learning with frozen language models},
  author={Tsimpoukelli, Maria and Menick, Jacob L and Cabi, Serkan and Eslami, SM and Vinyals, Oriol and Hill, Felix},
  journal={Advances in Neural Information Processing Systems},
  volume={34},
  pages={200--212},
  year={2021}
}

@article{liu2023visual,
  title={Visual instruction tuning},
  author={Liu, Haotian and Li, Chunyuan and Wu, Qingyang and Lee, Yong Jae},
  journal={Advances in neural information processing systems},
  volume={36},
  pages={34892--34916},
  year={2023}
}

@article{touvron2023llama,
  title={Llama: Open and efficient foundation language models},
  author={Touvron, Hugo and Lavril, Thibaut and Izacard, Gautier and Martinet, Xavier and Lachaux, Marie-Anne and Lacroix, Timoth{\'e}e and Rozi{\`e}re, Baptiste and Goyal, Naman and Hambro, Eric and Azhar, Faisal and others},
  journal={arXiv preprint arXiv:2302.13971},
  year={2023}
}

@article{yu2022coca,
  title={Coca: Contrastive captioners are image-text foundation models},
  author={Yu, Jiahui and Wang, Zirui and Vasudevan, Vijay and Yeung, Legg and Seyedhosseini, Mojtaba and Wu, Yonghui},
  journal={arXiv preprint arXiv:2205.01917},
  year={2022}
}

@inproceedings{nguyen2024open3dis,
  title={Open3dis: Open-vocabulary 3d instance segmentation with 2d mask guidance},
  author={Nguyen, Phuc and Ngo, Tuan Duc and Kalogerakis, Evangelos and Gan, Chuang and Tran, Anh and Pham, Cuong and Nguyen, Khoi},
  booktitle={Proceedings of the IEEE/CVF Conference on Computer Vision and Pattern Recognition},
  pages={4018--4028},
  year={2024}
}

@inproceedings{yan2024maskclustering,
  title={Maskclustering: View consensus based mask graph clustering for open-vocabulary 3d instance segmentation},
  author={Yan, Mi and Zhang, Jiazhao and Zhu, Yan and Wang, He},
  booktitle={Proceedings of the IEEE/CVF Conference on Computer Vision and Pattern Recognition},
  pages={28274--28284},
  year={2024}
}

@inproceedings{yin2024sai3d,
  title={Sai3d: Segment any instance in 3d scenes},
  author={Yin, Yingda and Liu, Yuzheng and Xiao, Yang and Cohen-Or, Daniel and Huang, Jingwei and Chen, Baoquan},
  booktitle={Proceedings of the IEEE/CVF Conference on Computer Vision and Pattern Recognition},
  pages={3292--3302},
  year={2024}
}

@article{sun2023eva,
  title={Eva-clip: Improved training techniques for clip at scale},
  author={Sun, Quan and Fang, Yuxin and Wu, Ledell and Wang, Xinlong and Cao, Yue},
  journal={arXiv preprint arXiv:2303.15389},
  year={2023}
}

@inproceedings{zhai2023sigmoid,
  title={Sigmoid loss for language image pre-training},
  author={Zhai, Xiaohua and Mustafa, Basil and Kolesnikov, Alexander and Beyer, Lucas},
  booktitle={Proceedings of the IEEE/CVF international conference on computer vision},
  pages={11975--11986},
  year={2023}
}

@article{tschannen2025siglip,
  title={Siglip 2: Multilingual vision-language encoders with improved semantic understanding, localization, and dense features},
  author={Tschannen, Michael and Gritsenko, Alexey and Wang, Xiao and Naeem, Muhammad Ferjad and Alabdulmohsin, Ibrahim and Parthasarathy, Nikhil and Evans, Talfan and Beyer, Lucas and Xia, Ye and Mustafa, Basil and others},
  journal={arXiv preprint arXiv:2502.14786},
  year={2025}
}

@inproceedings{kim2024ov,
  title={OV-MAP: Open-Vocabulary Zero-Shot 3D Instance Segmentation Map for Robots},
  author={Kim, Juno and Park, Yesol and Yoon, Hye-Jung and Zhang, Byoung-Tak},
  booktitle={2024 IEEE/RSJ International Conference on Intelligent Robots and Systems (IROS)},
  pages={13780--13786},
  year={2024},
  organization={IEEE}
}

@inproceedings{ngo2023isbnet,
  title={Isbnet: a 3d point cloud instance segmentation network with instance-aware sampling and box-aware dynamic convolution},
  author={Ngo, Tuan Duc and Hua, Binh-Son and Nguyen, Khoi},
  booktitle={Proceedings of the IEEE/CVF Conference on Computer Vision and Pattern Recognition},
  pages={13550--13559},
  year={2023}
}

@article{labbe2019rtab,
  title={RTAB-Map as an open-source lidar and visual simultaneous localization and mapping library for large-scale and long-term online operation},
  author={Labb{\'e}, Mathieu and Michaud, Fran{\c{c}}ois},
  journal={Journal of field robotics},
  volume={36},
  number={2},
  pages={416--446},
  year={2019},
  publisher={Wiley Online Library}
}

@article{bai2025qwen2,
  title={Qwen2.5-VL technical report},
  author={Bai, Shuai and Chen, Keqin and Liu, Xuejing and Wang, Jialin and Ge, Wenbin and Song, Sibo and Dang, Kai and Wang, Peng and Wang, Shijie and Tang, Jun and others},
  journal={arXiv preprint arXiv:2502.13923},
  year={2025}
}

@inproceedings{xu2025sampro3d,
  title={SAMPro3D: Locating SAM Prompts in 3D for Zero-Shot Instance Segmentation},
  author={Xu, Mutian and Yin, Xingyilang and Qiu, Lingteng and Liu, Yang and Tong, Xin and Han, Xiaoguang},
  booktitle={2025 International Conference on 3D Vision (3DV)},
  pages={1222--1232},
  year={2025},
  organization={IEEE}
}

@inproceedings{huang2024openins3d,
  title={Openins3d: Snap and lookup for 3d open-vocabulary instance segmentation},
  author={Huang, Zhening and Wu, Xiaoyang and Chen, Xi and Zhao, Hengshuang and Zhu, Lei and Lasenby, Joan},
  booktitle={European Conference on Computer Vision},
  pages={169--185},
  year={2024},
  organization={Springer}
}

@inproceedings{rasheed2024glamm,
  title={Glamm: Pixel grounding large multimodal model},
  author={Rasheed, Hanoona and Maaz, Muhammad and Shaji, Sahal and Shaker, Abdelrahman and Khan, Salman and Cholakkal, Hisham and Anwer, Rao M and Xing, Eric and Yang, Ming-Hsuan and Khan, Fahad S},
  booktitle={Proceedings of the IEEE/CVF Conference on Computer Vision and Pattern Recognition},
  pages={13009--13018},
  year={2024}
}

@inproceedings{guo2024regiongpt,
  title={Regiongpt: Towards region understanding vision language model},
  author={Guo, Qiushan and De Mello, Shalini and Yin, Hongxu and Byeon, Wonmin and Cheung, Ka Chun and Yu, Yizhou and Luo, Ping and Liu, Sifei},
  booktitle={Proceedings of the IEEE/CVF Conference on Computer Vision and Pattern Recognition},
  pages={13796--13806},
  year={2024}
}

@article{cheng2024spatialrgpt,
  title={Spatialrgpt: Grounded spatial reasoning in vision-language models},
  author={Cheng, An-Chieh and Yin, Hongxu and Fu, Yang and Guo, Qiushan and Yang, Ruihan and Kautz, Jan and Wang, Xiaolong and Liu, Sifei},
  journal={Advances in Neural Information Processing Systems},
  volume={37},
  pages={135062--135093},
  year={2024}
}

@article{bai2023qwen,
  title={Qwen technical report},
  author={Bai, Jinze and Bai, Shuai and Chu, Yunfei and Cui, Zeyu and Dang, Kai and Deng, Xiaodong and Fan, Yang and Ge, Wenbin and Han, Yu and Huang, Fei and others},
  journal={arXiv preprint arXiv:2309.16609},
  year={2023}
}

@inproceedings{lu2023ovir,
  title={Ovir-3d: Open-vocabulary 3d instance retrieval without training on 3d data},
  author={Lu, Shiyang and Chang, Haonan and Jing, Eric Pu and Boularias, Abdeslam and Bekris, Kostas},
  booktitle={Conference on Robot Learning},
  pages={1610--1620},
  year={2023},
  organization={PMLR}
}

@article{boudjoghra2024openyolo3d,
  title={Open-yolo 3d: Towards fast and accurate open-vocabulary 3d instance segmentation},
  author={Boudjoghra, Mohamed El Amine and Dai, Angela and Lahoud, Jean and Cholakkal, Hisham and Anwer, Rao Muhammad and Khan, Salman and Khan, Fahad Shahbaz},
  journal={arXiv preprint arXiv:2406.02548},
  year={2024}
}

@article{ren2024grounded,
  title={Grounded sam: Assembling open-world models for diverse visual tasks},
  author={Ren, Tianhe and Liu, Shilong and Zeng, Ailing and Lin, Jing and Li, Kunchang and Cao, He and Chen, Jiayu and Huang, Xinyu and Chen, Yukang and Yan, Feng and others},
  journal={arXiv preprint arXiv:2401.14159},
  year={2024}
}

@inproceedings{cheng2024yoloworld,
  title={Yolo-world: Real-time open-vocabulary object detection},
  author={Cheng, Tianheng and Song, Lin and Ge, Yixiao and Liu, Wenyu and Wang, Xinggang and Shan, Ying},
  booktitle={Proceedings of the IEEE/CVF conference on computer vision and pattern recognition},
  pages={16901--16911},
  year={2024}
}
\end{document}